%% file: main.tex
\documentclass[letterpaper]{article} 
\usepackage{aaai2026}  
\usepackage{iftex}
\ifPDFTeX
\usepackage{times}  
\usepackage{helvet}  
\usepackage{courier}
\fi
\usepackage[hyphens]{url}  
\usepackage{graphicx} 
\usepackage{natbib}  
\usepackage{caption} 
\usepackage{float}
\usepackage{newfloat}
 \usepackage{amssymb}
\usepackage{listings}
\usepackage{booktabs}
\usepackage{adjustbox}
\usepackage{multirow}
\usepackage{xcolor}
\usepackage{amsmath}
\usepackage{amsthm} 
\usepackage{tabularx} 
\usepackage{array}
\usepackage{nicematrix} 
\newcolumntype{Y}{>{\raggedright\arraybackslash}X}

\DeclareCaptionStyle{ruled}{labelfont=normalfont,labelsep=colon,strut=off} 
\floatstyle{ruled}
\newfloat{listing}{tb}{lst}{}
\floatname{listing}{Listing}
\ifPDFTeX
\fi
\title{From Black Box to Controller: Inference-Time Steering of LLM via Feature Amplification}
\author{
    Keqi Chen,
    Zekai Sun, 
    Yuhua Wen, 
    Tianyi Tian,
    Xinyang Huang,
    Huijun Lian, \\ 
    Yingming Gao, 
    Ya Li
}
\affiliations{
    \textsuperscript{}Beijing University of Posts and Telecommunications \\
    Beijing, China \\
        {ckqqqq,sunzekai,yuhuawen,tian\_tianyi,xinyanghuang,lianhuijunlybl,yingming.gao,yli01}@bupt.edu.cn
}

\usepackage{bibentry}

\begin{document}

\maketitle

\begin{abstract}

\input{sec/0abstract}
\end{abstract}
\section{Introduction}
\input{sec/1intro}

\section{Related Work}
\begin{figure*}[h!]
    \centering
    \includegraphics[width=0.95\linewidth]{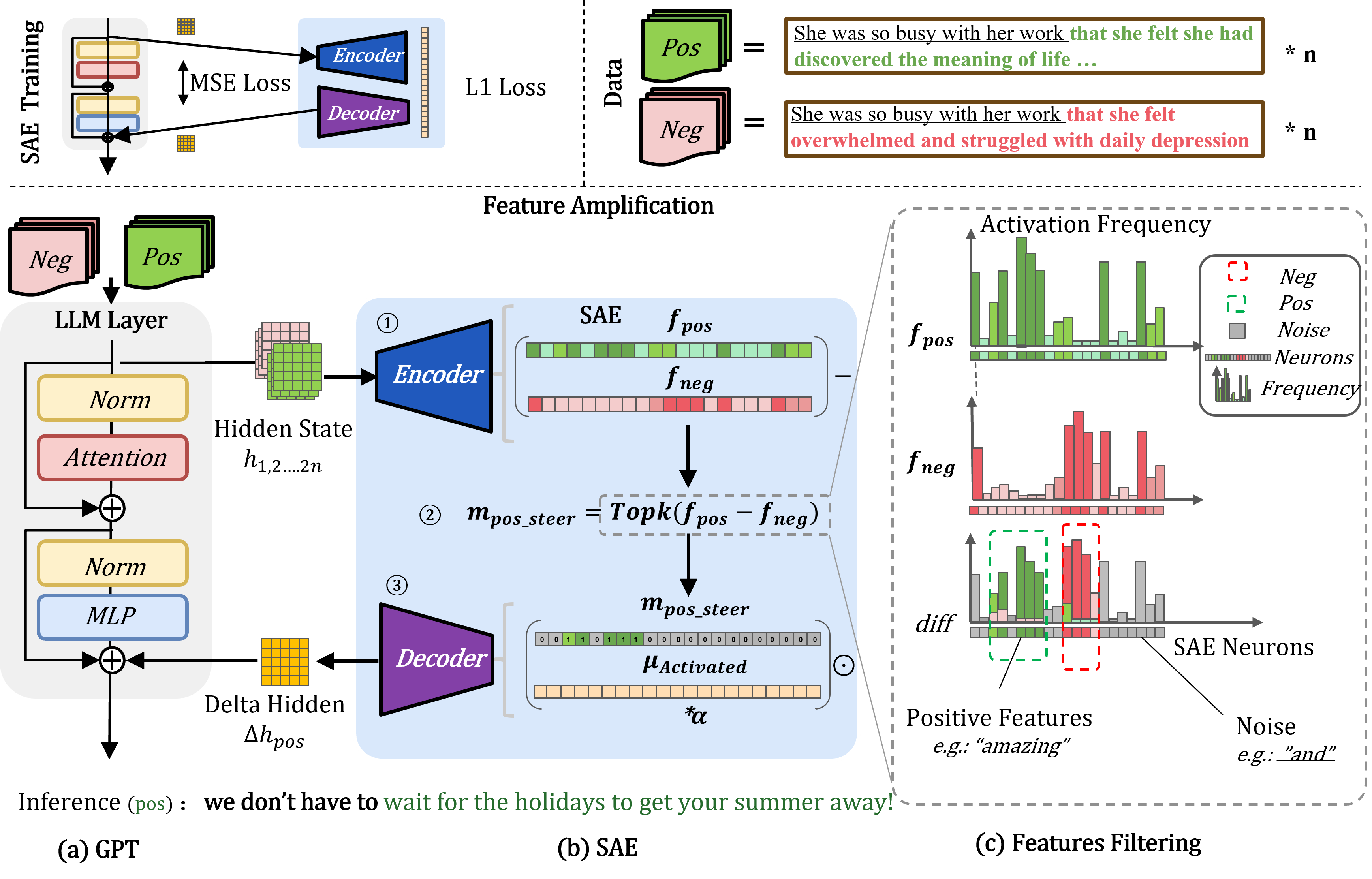}
    \caption{An overview of the NeuroSteer architecture. The feature amplification includes steps: \textcircled{1} \textbf{SAE Feature Collections}: The SAE collects neuron activation frequencies of positive and negative samples; \textcircled{2} \textbf{Feature Filtering}: The SAE extracts target neurons and denoises based on frequency differences; \textcircled{3} \textbf{Inference-time Feature Amplifying}: The SAE activates target neuron combinations and decodes their effects into the GPT hidden space, which is also shown in Figure~\ref{fig:data-construction}. }
    \label{fig: main_picture}
  \end{figure*}
\label{sec:related_work}
\input{sec/2related}

\section{Method}
\label{method}
\input{sec/3method}
\section{Experiment}
\label{experiment}
\input{sec/4experiment}
\section{Conclusion}
\input{sec/5conclusion}
\bibliography{aaai2026}

\appendix
\input{sup}
\end{document}

%% file: sec/0abstract.tex
Inference-Time steering is crucial for deploying Large Language Models (LLMs) in real-world applications, where they must flexibly adjust behavior preference. For instance, the model should dynamically adjust its caution level based on the risk of the scenario. This requirement introduces a challenge: modulating the intensity of LLM preferences. Prompt engineering can require task-specific prompt design, while fine-tuning requires additional optimization and data. These requirements can make repeated behavioral adjustments costly. To address this, we introduce \textbf{NeuroSteer}, a plug-and-play framework that enables flexible, quantitative adjustment of LLM preferences at inference time. NeuroSteer employs a Sparse Autoencoder (SAE) as its controller. Comparing neuron activation frequencies between positive and negative examples isolates “feature neurons” that correspond to target preferences such as politeness and positive sentiment. During inference, the activations of selected SAE features are amplified in the latent representations to steer the output distribution. Users can adjust the strength of this steering on a per-scenario or per-user basis, granting the LLM fine-grained, quantifiable trait modulation. Our experiments show that NeuroSteer can alter sentiment, toxicity, stance, and politeness at inference time. It achieves higher target-direction steering scores than the compared baselines in several settings, with task- and strength-dependent trade-offs in generation quality. The anonymous demo is accessible.\footnote{\url{http://aaai.anonymous-demo.fun}}.

%% file: sec/1intro.tex
As LLMs are deployed widely, there is an increasing expectation for LLMs to adapt their preferences across various scenarios. For instance, LLMs should exercise greater caution when answering financial questions compared to open-domain conversations. Such adjustments are often dynamic. For instance, LLMs are required to increase their 'caution level' to refuse risk-sensitive questions in some cases. But excessive caution leads to LLMs' unnecessary refusals and over-refusal, which can result in poor user experience. Consequently, scenario-aware adjustments of LLMs' caution levels are crucial. Similarly, adaptive tuning of other trait dimensions - such as toxicity control and politeness adjustment - proves critical for task-specific adaptation.

Current solutions for such adaptive tuning include full fine-tuning~\cite{devlin2018bert}, LoRA~\cite{hu2021lora}, prompting, and RLHF~\cite{ouyang2022traininglanguagemodelsfollow}. However, these methods require substantial resources, including prompt adjustments, datasets, retraining for specific tasks, etc., which makes them less flexible. These approaches do not by themselves provide a common scalar control for adjusting the strength of a target behavior at inference time.

  \begin{figure}[t!]
    \centering
\includegraphics[width=0.47\textwidth]{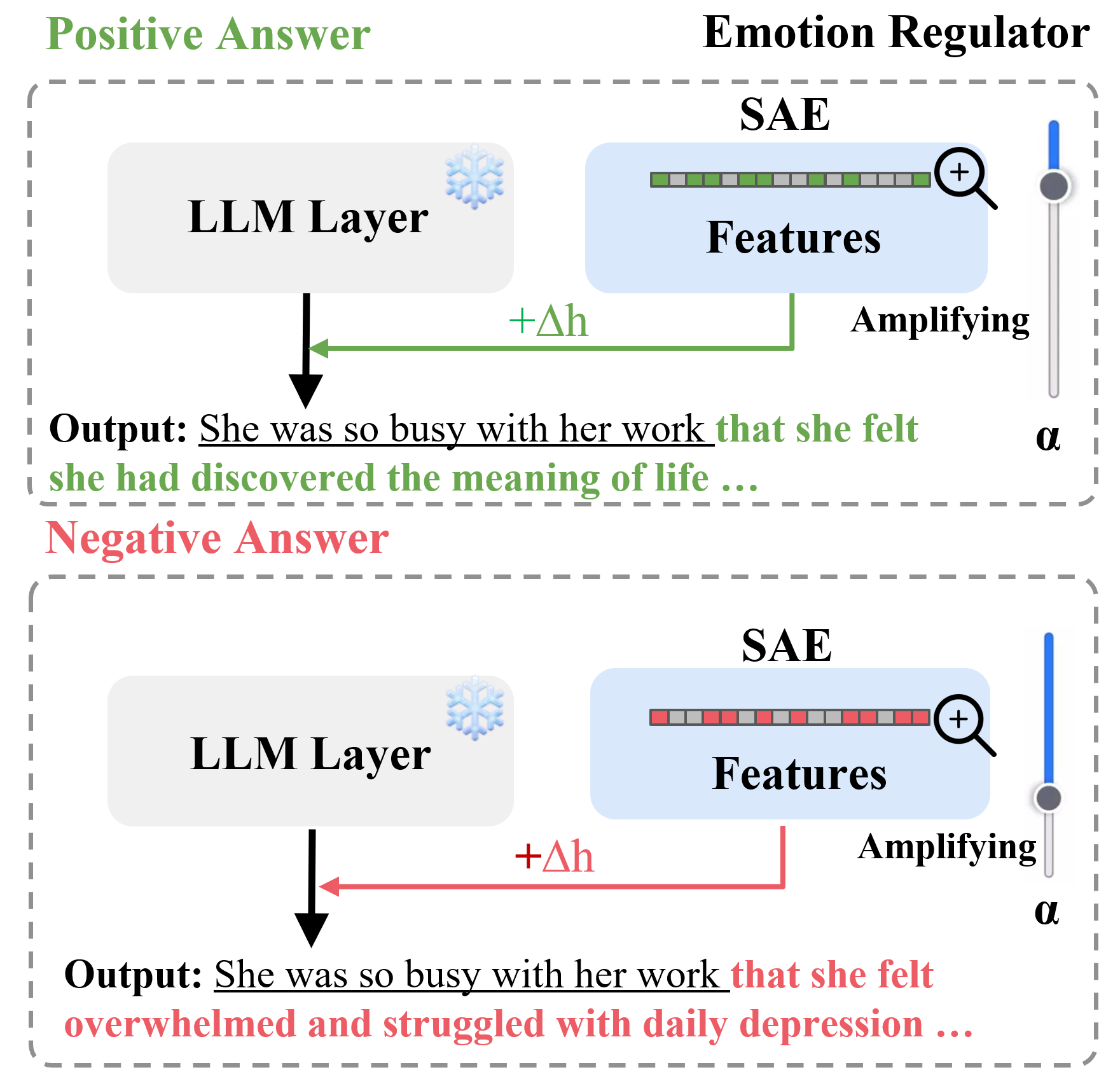}
    \centering
    \caption{Overview of inference-time steering. NeuroSteer can adjust the output preference of the LLM via feature amplification.}
    \label{fig:data-construction}
  \end{figure}
  \definecolor{darkgreen}{rgb}{0.0, 0.5, 0.0} 
  \definecolor{darkred}{rgb}{0.5, 0.0, 0.0} 

  \definecolor{darkgreen}{rgb}{0.0, 0.5, 0.0} 
  \definecolor{darkred}{rgb}{0.5, 0.0, 0.0} 

Activation engineering provides new solutions. Studies~\cite{jiang2024originslinearrepresentationslarge,linearrepresentation,linearrepresentation2} show that LM encodes knowledge along linear directions in hidden activations, enabling latent edits without weight changes. To adjust model output, activation engineering methods~\cite{elhage2022superposition,li2024inference, panickssery2024steeringllama2contrastive} use the steering vector to shift LM activation to a linear direction to change LM output with some promise, but they also raise a question: Do LLMs possess representations of traits like 'emotion' and 'politeness'? If so, could we achieve quantitative control over an LLM's trait preferences by amplifying or suppressing representations?

Previous Studies~\cite{gurnee2024languagemodelsrepresentspace} show that LLM activation does not have a single linear dimension to represent traits. The trait is composed of multiple low-dimensional features, such as simple logic and tokens. The combination of these low-dimensional vectors consists of abstract and high-level traits, such as answer rejection or positive emotion.

SAE provides a solution capable of capturing and intervening in these low-dimensional features. The SAE is an unsupervised method with the original goal of discovering interpretable features. It is trained by reconstructing the activation vectors of the LLM, which can break down high-dimensional and coupling activations into more fundamental, understandable units. Sparse features provide candidate directions for more targeted interventions. However, sparsity does not guarantee feature independence, and amplifying a feature can still have off-target effects. This allows it to serve as an "amplifier" of LLM features.

Building upon the above exploration, we propose NeuroSteer, which modulates the degree of targeted behavior preferences in LLMs at inference time. We use SAE as LM's controller to regulate these features. As shown in Figure~\ref{fig:data-construction}, the SAE learns to map the high-dimensional features into sparse and basic features. To steer the LM toward a target trait, we activate and combine SAE neurons corresponding to basic target features. By amplifying or suppressing these features, LLMs can achieve macro-regulation of output preference. During inference, the residual addition of these features adjusts the probability distribution on LLMs’ next-token predictions, steering the LM output toward the desired trait.

To select SAE neurons representing target features, we use a frequency filter on comparative data pairs. Neurons that are frequently activated in positive samples but not in negative ones are key to positive (target) traits. Neurons activated in both samples represent noise and are not activated for steering.

  Our contributions can be summarized as follows:

We develop NeuroSteer, an LLM control framework that introduces the SAE features regulation for inference-time steering. We evaluate NeuroSteer on sentiment, toxicity, stance, and politeness steering tasks. It improves several steering metrics relative to the compared baselines, while exposing trade-offs between steering strength and generation quality. Deeper insights are gained into how concepts are represented in LM by layer, and how amplifying and combining basic features affects the LM’s behavior.







%% file: sec/2related.tex
The work of NeuroSteer is inspired by previous work on activation engineering and sparse autoencoders.
\subsection{Activation Engineering}
Although LM's activations are often seen as black boxes, studies show they develop semantically meaningful internal structures .~\citet {jiang2024originslinearrepresentationslarge} and~\citet{li2024inference} suggest that semantic concepts are generally encoded linearly in the hidden representation space of LMs. For instance, concepts such as sentiment polarity~\cite{tigges2023linearrepresentationssentimentlarge}, time, and space~\cite{gurnee2024languagemodelsrepresentspace} can be represented through linear directions in hidden states.

Following the idea of linear representations, approaches like~\citet{panickssery2024steeringllama2contrastive} compute mean hidden state difference activations along the vector connecting the means of the true and false distributions. As shown in Table~\ref{table:compare}, these methods employ the mean difference vector to improve the truthfulness of LM. Similarly, methods such as LM-steer~\cite{dong2023steerlm,tigges2023linearrepresentationssentimentlarge} steer the LM latent state with steering vectors using SVD decomposition. Most activation engineering methods, such as ITI and CAA~\cite{ITI-2,yang2025enhancingsemanticconsistencylarge,stickland202steeringwithoutsideeffect} calculate mean differences between positive and negative samples, applying a single steering vector directly on LM hidden states.
Despite their potential, activation engineering inside LLM faces two critical challenges rooted in the fundamental properties of LM representations. (1) A single steering vector restricts the scope and capability of controlling complex LM behavior. (2) The dense, highly superimposed nature of LM hidden layers causes unintended noise during interventions, and calculating the mean difference vector introduces more noise, leading to undesired outputs. Direct activation editing can therefore introduce off-target effects.  As shown by~\citet{elhage2022superposition}, individual neurons rarely encode single features – GPT-2's 768-dimensional neurons must encode over $10^{5}$ distinct features~\cite{gurnee2024universalneuronsgpt2language}.  This indicates that activating a single LLM neuron can affect a large number of distinct features in its representation, which introduces noise~\cite{stickland202steeringwithoutsideeffect}. The mean single difference vector approach limits intervention flexibility. Representing composite behaviors with a single direction may also limit intervention flexibility.
\input{sec/1table0}

  \subsection{Sparse Autoencoders}
SAE offers potential for enabling finer-grained, controllable, and interpretable intervention methods. SAEs learn by reconstructing activation vectors between LLM layers through autoencoders, projecting the dense internal activations of LLMs into a sparse representational space to achieve disentanglement~\cite{templeton2024scaling}. SAE imposes sparsity constraints on high-dimensional latent spaces (typically 10-100 times larger than LLM embeddings). This encourages sparse activation patterns that can correspond to distinct semantic features, without guaranteeing statistical independence. For example, in GPT-2 SAEs, specific neurons fire exclusively for concepts like 'email addresses' or 'chemical formulas'~\cite{gpt2_saes}. The SAE architecture naturally separates superimposed features: Given an encoded representation on LM, the encoder produces sparse features, where only a small subset of neurons are activated, while the decoder reconstructs the original encoded representation. The overcomplete basis provides candidate features for targeted interventions. Feature overlap and unintended effects can nevertheless remain and must be evaluated empirically.

%% file: sec/1table0.tex
\begin{table}[h!]
    \centering
    \renewcommand{\arraystretch}{1.2} 
    \adjustbox{max width=\linewidth}{%
    \begin{tabular}{@{}lccccc@{}}
        \toprule
        \textbf{Methods} & \rotatebox{0}{\textbf{Quantifiable}} & \rotatebox{0}{\textbf{Plug-and-Play}} & \rotatebox{0}{\textbf{Infer Delay}} & \rotatebox{0}{\textbf{Linearity}} & \rotatebox{0}{\textbf{Whitebox}} \\
        \midrule
        Fine-Tuning & \textcolor{darkred}{$\times$} & \textcolor{darkred}{$\times$} & \textcolor{darkgreen}{$\checkmark$} & \textcolor{darkred}{$\times$} & \textcolor{darkred}{$\times$} \\
        LoRA (\citeyear{hu2021lora}) & \textcolor{darkred}{$\times$} & \textcolor{darkred}{$\times$} & \textcolor{darkgreen}{$\checkmark$} & \textcolor{darkred}{$\times$} & \textcolor{darkred}{$\times$} \\
        ITI (\citeyear{li2024inference}) & \textcolor{darkgreen}{$\checkmark$} & \textcolor{darkgreen}{$\checkmark$} & \textcolor{darkgreen}{$\checkmark$} & \textcolor{darkred}{$\times$} & \textcolor{darkred}{$\times$} \\
        CAA (\citeyear{panickssery2024steeringllama2contrastive}) & \textcolor{darkgreen}{$\checkmark$} & \textcolor{darkgreen}{$\checkmark$} & \textcolor{darkgreen}{$\checkmark$} & \textcolor{darkred}{$\times$} & \textcolor{darkred}{$\times$} \\
        LM-Steer (\citeyear{han2024word}) & \textcolor{darkgreen}{$\checkmark$} & \textcolor{darkred}{$\times$} & \textcolor{darkred}{$\times$} & \textcolor{darkred}{$\times$} & \textcolor{darkred}{$\times$} \\
        \textbf{NeuroSteer} & \textcolor{darkgreen}{$\checkmark$} & \textcolor{darkgreen}{$\checkmark$} & \textcolor{darkgreen}{$\checkmark$} & \textcolor{darkgreen}{$\checkmark$} & \textcolor{darkgreen}{$\checkmark$} \\
        \bottomrule
    \end{tabular}
    }
    \caption{Comparison of steering methods. NeuroSteer demonstrates advantages in flexibility and application potential.}
    \label{table:compare}
\end{table}

%% file: sec/3method.tex
NeuroSteer introduces the SAE as a controller. During forward propagation, the SAE amplifies a set of selected sparse features, thereby applying a preference steering effect on the LLM's output.
  \subsection{Method Overview}

  As illustrated in Figure~\ref{fig: main_picture}, NeuroSteer controls LLM preference degree through trained sparse autoencoders (SAEs) and target positive/negative data sampling. This method first requires SAE pretraining. Training on any pretrained corpora like OpenWebText enables SAEs to learn mappings between token activations and fundamental features. Building on pretrained SAEs, we then perform forward propagation to filter target features for specific behavior preference. In detail, we activate the LLM with positive/negative data pairs, recording SAE activation frequencies, computing inter-sample neuronal activation differences, and ultimately selecting target features. During inference, NeuroSteer amplifies these target features by a certain ratio to achieve real-time steering of the generation process and modulation of model preferences.

\subsection{SAE Training}
The SAE is an unsupervised autoencoder method. It encodes the residual activations of each layer in the LLM using an autoencoder, mapping them to a high-dimensional sparse space, and then reconstructs the activation vectors through the decoder. The SAE has two losses: \textbf{MSE loss} measures the difference between the reconstructed activation and the original activation, and the other is the \textbf{L1 loss} for SAE sparsity. We follow the training setting of previous works~\cite{gpt2_saes,templeton2024scaling} to pretrain the SAE. Once the SAE is trained, quantifiable adjustments for each specific task can be achieved in a high-dimensional sparse space through forward propagation, without requiring any gradient updates.

\subsection{Data Pairs}
NeuroSteer's feature collection relies on a small batch of positive-negative sample pairs. As illustrated in the figures~\ref{fig: main_picture}, the sentiment steering task employs opposite text pairs which are easily accessible in classification datasets. Activation sampling of positive and negative examples captures feature distinctions between target preferences. Such data suit the modulation of base models lacking prompt-following capabilities such as gpt-2.

Data pair samples can also be obtained through prompt-mask sampling. For instance, the common response to 'Please tell me about Microsoft's revenue' differs from the cautious version 'Please tell me about Microsoft's revenue, but don't provide sensitive data on private dataset.' By sampling answer activations from these paired prompts, the SAE collects steering target feature differentiae for adjusting the degree of caution.

\subsection{SAE Features Collection }
With trained SAE, we select target SAE features by forward processing through statistical filtering on small dataset.

\noindent\textbf{SAE Encoding:} First, we input the positive and negative texts into the GPT model and collect the activation by forward propagation. For positive sentiment texts, $n_{\text{pos}}$ tokens are used, and for negative counterparts, $n_{\text{neg}}$ tokens are used. For each token hidden state $\mathbf{h}_{\text{GPT}}^{i} \in \mathbb{R}^{d}$, the SAE encoder maps it to a sparse representation in higher-dimensional space :
  \begin{equation}
    \label{eq:encoding}\mathbf{h}_{\text{SAE}}^{i}= \mathrm{Enc}\left( \mathbf{h}
    _{\text{GPT}}^{i}\right), \quad \mathbb{R}^{d}\rightarrow \mathbb{R}^{k},
  \end{equation}
  where $k \gg d$ enables sparse feature separation through the SAE's dimensional expansion.

  \noindent\textbf{Frequency Statistics:} For each SAE neuron $j \in \{1,...,k\}$, we compute activation frequencies across positive$f_{\text{pos},j}$ and negative samples $f_{\text{neg},j}$:
  \begin{equation}
    \label{eq:stat}f_{\text{pos},j}= \frac{1}{n_{\text{pos}}}\sum_{i=1}^{n_{\text{pos}}}
    \mathbb{I}(\mathbf{h}_{\text{SAE},j}^{i} > 0) \in [0,1],
  \end{equation}
  where $\mathbb{I}(\cdot)$ denotes the indicator function. The negative-sample frequency is defined analogously. Collecting all $k$ scalar frequencies gives vectors $\mathbf{f}_{\mathrm{pos}},\mathbf{f}_{\mathrm{neg}}\in[0,1]^k$.

  \subsection{Feature Filtering}

  Figure~\ref{fig: main_picture}(c) shows that green neurons activate for positive samples, while red neurons represent activations for negative samples, with color intensity indicating activation frequency. Neurons encoding positive features exhibit higher activation frequency in positive samples. To filter out the noise and select the TopK neurons for target steering, the difference in activation frequencies between positive and negative samples is computed, as illustrated in Equation~\ref{eq:topk}.

  \begin{align}
    \label{eq:topk}\mathbf{m}_{\mathrm{pos\_steer}}= \operatorname{TopKMask}(\mathbf{f}_{\mathrm{pos}}-\mathbf{f}_{\mathrm{neg}}) \in \{0,1\}^{k},
  \end{align}
  Here, $\operatorname{TopKMask}$ sets entries corresponding to the $K$ largest activation-frequency differences to one and all other entries to zero. Selection is performed across features; $m_{\mathrm{pos\_steer},j}$ is the scalar mask entry for feature $j$.
  \subsection{Inference-time Feature Amplifying}
We construct a latent steering vector using the feature mask and the vector of mean feature activations $\boldsymbol{\mu}\in\mathbb{R}^k$. The operation is an elementwise product, denoted by $\odot$, rather than a dot product.

\noindent\textbf{SAE Decoding and Residual Addition}: We normalize the masked latent vector and scale it using the steering coefficient $\alpha$ before decoding:

\begin{align}
    \Delta\mathbf{z}= \mathbf{m}_{\mathrm{pos\_steer}}\odot \boldsymbol{\mu} \in \mathbb{R}^{k},
\end{align}

The latent vector remains $k$-dimensional until decoding maps it back to the $d$-dimensional model residual space. Varying $\alpha$ controls the intervention strength, although the resulting behavioral change need not be linear or monotonic:
\begin{equation}
    h_{\mathrm{GPT}}^{'} = h_{\mathrm{GPT}} + \mathrm{Dec}(\alpha \cdot \frac{\Delta\mathbf{z}}{\sigma(\Delta\mathbf{z}) + \epsilon}) \in \mathbb{R}^{d},
    \label{eq:nb}
\end{equation}

\subsection{Transfer to Other Tasks:}
In conclusion, this section demonstrates NeuroSteer's output steering capability through emotional polarity reversal in LLM, transforming positive sentiment to negative. The steering mechanism extends to toxicity mitigation via data pairs of toxic and non-toxic samples. Further generalization enables stance modification and politeness adjustment, as empirically validated in our experiments.

%% file: sec/4experiment.tex
\input{sec/4table1}

\input{sec/4table2}

The SAE learns and pre-trains based on fundamental features, making it a cross-task controller. Task-specific feature selection uses labeled examples without updating the base model weights. We evaluate four steering tasks: \textbf{Sentiment Steering}, \textbf{Detoxify}, \textbf{Stance Steering}, and \textbf{Politeness Steering}.

All experiments follow similar data and evaluation frameworks: first, extracting the former part of the prompt with different polarities. These prompts tend to induce corresponding polarized outputs. For instance, when given a prompt with negative sentiment in the first sentence, the model is more likely to generate a continuation with similarly negative sentiment in the latter sentence. After generating the target polarity steering texts, a classifier is used to classify the polarity of all generated texts. The proportion of generated text with the target polarity steering is then calculated. This proportion serves as a measure of the steering capability of the method, with a higher change indicating stronger steering ability.

\subsection{SAE Training}

We use GPT-2-small~\cite{radford2019language} and Gemma-2-2B as base models. The main GPT-2 experiments intervene at layer 6; the Gemma experiments use the layer-12 SAE described below.
We train the SAE on OpenWebText for GPT-2, and load the gemma-2-2b (Layer=12) SAE on gemmascope~\cite{lieberum2024gemmascopeopensparse}. Our SAE is trained with batch 64, an L1 coefficient of 8e-5, and a learning rate of 4e-4 for 300 million tokens on OpenWebText. 
GPT-2 SAE training took approximately 30 hours on RTX 4090 hardware. The reported dead-feature fraction was 28\%. The SAE weights will be open-sourced on Hugging Face.

\subsection{Inference-time Steering On Four Tasks}
The sentiment steering and detoxification tasks have provided baselines, making them suitable for cross-method comparison. The stance and politeness steering tasks are innovative tasks that have been scarcely explored in previous papers. So we reproduce baseline results for our method on these two tasks.

\noindent\textbf{Sentiment Steering:} In this task, we calculate and select the steering neurons on 3452 data samples from Stanford Sentiment Treebank (SST5, \citealp{socher2013recursive}) for sentiment steering. SST5 samples are classified for positive or negative sentiment and are used to compute positive and negative frequency differences for targeted neuron selection. We assess the model for sentiment evaluation using the multiclass sentiment analysis dataset~\citep{shrivastava_sentiment_analysis_nodate}, an out-of-domain dataset with 2500 prompts from an Open-Web sentiment classifier~\citep{hartmann2023}.

\noindent\textbf{Detoxify:} By controlling the toxic generation, Neurosteer can reduce harmful or offensive content. Jigsaw toxicity classification dataset~\cite{jigsaw-toxic-comment-classification-challenge} with 155k data samples is a very suitable dataset for calculating $\Delta h$. For out-of-domain evaluation, we use a random sample of 10K non-toxic prompts from the RealToxicity Prompts dataset~\citep{gehman2020realtoxicityprompts} to assess toxicity reduction in GPT-2 generation.

\noindent\textbf{Stance Steering:} Stance detection determines whether the speaker is in favor of, against, or neutral to a target event. After steering, GPT can reverse the originally agreed-upon stance in the latter part of a sentence. We test stance reversal on the StanceSentences dataset~\cite{reyesexplainable}, which contains 972 samples, for feature-vector construction and evaluation.

\noindent\textbf{Politeness Steering:} Controlling politeness in language models ensures that the generated outputs align with expected social norms and etiquette. The politeness control experiment follows the same approach as stance control. In the politeness steering experiment, we use HuggingFace politeness corpus (\citeyear{politeness-corpus}) with 10.9k samples for training and testing, with a 3:1 train-test split.

\subsection{Steering Baselines}

Trained methods include \textbf{DAPT}~\cite{gururangan2020don} and \textbf{PromptT5}~\cite{raffel2020exploring}, which pretrain the LM on a non-toxic subset of OpenWebText (filtered via the Perspective API), and \textbf{PPLM}~\cite{dathathri2019plug}, which uses gradients from sentiment and toxicity classifiers to update the LM's hidden representations. Additionally, \textbf{LM-Steer}~\cite{han2024word} freezes the GPT layers and adds a fine-tuning layer.

Plug-and-Play methods include \textbf{GeDi}~\cite{krause2020gedi}, which uses bayesian rules for class-conditioned generation; \textbf{MuCoLa}~\cite{kumar2022gradient}, which formulates text generation as an optimization problem based on classifier scores; and \textbf{DExperts}~\cite{liu-etal-2021-dexperts}, which offsets the LM logits using the difference between positive and negative label classifier scores.

Activation engineering method includes \textbf{LM Steering}, which uses the mean single difference vector to steer LLM on target behaviors.

Due to the lack of relevant baselines on stance and politeness experiments, we provided GPT-2 with full \textbf{fine-tuning} and \textbf{LoRA} as baselines in stance and politeness steering experiments.

\subsection{Evaluation metrics}
\label{metric:5.1.2} Effective steering improves steering metrics compared to uncontrolled generation while maintaining text quality metrics.

\noindent\textbf{Steering Metrics:} The steering efficiency of the method is measured by the proportion of target polarity texts. Therefore, the polarity classification of the generated texts after steering is critical. In the sentiment steering task, we use an outstanding sentiment classifier~\cite{hartmann2023} for sentiment classification. In the detoxification task, the Google Perspective API is employed to assess the toxicity polarity of the controlled generations. For the innovative tasks, we carefully design two prompts to instruct the Qwen-based evaluator~\cite{bai2023qwen} to classify the generated text for stance and politeness polarity.

\noindent\textbf{Text Quality:} Steering the model towards specific outputs can degrade the text quality. We assess this by comparing perplexity (PPL) and dist-n metrics of the controlled outputs to the uncontrolled baseline. Lower PPL indicates greater likelihood under the evaluation model, while higher dist-n indicates greater lexical diversity. These metrics capture fluency and lexical diversity, but do not by themselves establish factual accuracy, coherence, or task success.

\subsection{Results Evaluation}
  \input{sec/4table3}

Tables~\ref{table:sentiment} and~\ref{table:toxic} show improvements on several steering metrics, with trade-offs that depend on the task and steering strength. For positive sentiment steering, NeuroSteer reaches 99.19\% positivity on negative prompts, compared with the strongest baseline score of 57.26\%, an improvement of 41.93 percentage points. On neutral prompts, the corresponding best scores are 99.20\% and 94.57\%, a difference of 4.63 percentage points. These maxima occur at $\alpha=2.1$, where output PPL is 125.80, compared with 29.28 for the original GPT-2; stronger steering therefore does not uniformly preserve fluency.

For detoxification, NeuroSteer achieves a maximum-toxicity score of 0.069, compared with the lowest baseline score of 0.249, an absolute reduction of 0.180. Its toxicity probability is 0.008, compared with the lowest baseline value of 0.088, a reduction of 0.080. The compared methods use different backbone sizes, so these results should not be interpreted as a controlled comparison of the steering mechanism alone.

Table~\ref{table:stance and polite} also shows direction-dependent outcomes. At $\alpha=0.3$, NeuroSteer reaches 66.0\% in the agree-to-disagree column, compared with 44.0\% for full fine-tuning, but reaches 69.0\% in the reverse direction, below full fine-tuning at 76.0\%. For politeness, NeuroSteer at the same strength reaches the highest reported steering scores in both directions, while its PPL rises to 34.82, compared with 8.34 for LoRA. These results support adjustable steering rather than uniform superiority across metrics. Table~\ref{gentext} provides illustrative generations at different intervention strengths.

  \subsection{Computing Efficiency}
  \label{sec:efficiency}

  In the reported timing setup, NeuroSteer required less task-specific preparation time than LoRA. In tests, NeuroSteer (CPU-only, Intel Xeon) processed 3,000 samples in 14\% of the time required by LoRA (GPU-accelerated, NVIDIA RTX 4090). It achieves comparable performance using 600 samples, reported as 30\% of the data used in the corresponding comparison. Because the methods use different hardware and preparation procedures, this observation does not establish a hardware-matched speedup or an end-to-end inference latency improvement. The pretrained SAE also has an upfront training cost.
 \subsection{Ablation Steering}
 \begin{figure}[ht]
    \centering
    \includegraphics[width=\linewidth]{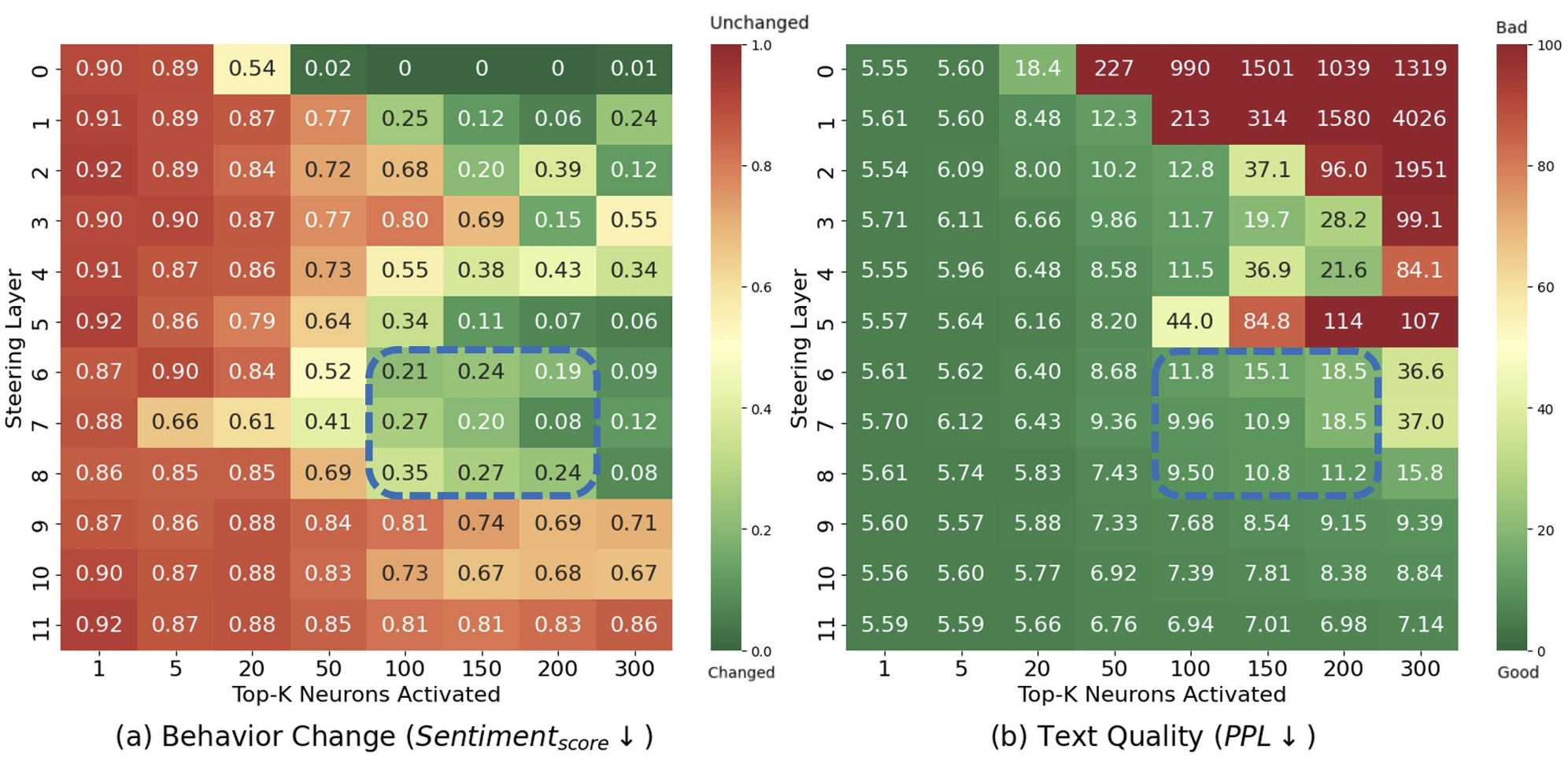}
    \caption{The overview of the steering experiment across layers. (a) shows that greener outputs correspond to better sentiment reversal performance, while (b) demonstrates that greener results maintain higher text quality. The parameter values inside the dashed-line box balance behavior change and text quality preservation.}
    \label{fig:layer_search}
  \end{figure}

  Since different layers of LM represent distinct semantic information, the choice of layer to steer impacts the intervention effect. To understand the semantic representation across LM layers, we expanded the grid search in the sentiment control experiment to examine the behavior of SAE neurons across different GPT layers. We varied both the layer (0–11) and the Top-K value of target SAE neurons, sorting them based on activation frequency differences (K), with higher activity in the target data and lower activity in the source data.

  Figure~\ref{fig:layer_search} shows layer-dependent steering behavior across the 12 GPT-2 layers. Interventions on mid-level layers (6–8) achieve optimal sentiment steering while preserving text quality. Surface layers (0–2) lead to repetitive negative tokens (e.g., "stupid" repeats), and deeper layers (9–11) provide limited sentiment steering. These observations characterize the tested interventions; they do not establish that deeper layers lack sentiment representations.

  The dashed box in Figure~\ref{fig:layer_search} highlights activating roughly the top 150 neurons in layers 6-8, encoding most relatively sentiment features via 100–350. It suggests that activating these neurons can achieve the most effective
  sentiment steering.

%% file: sec/4table1.tex
\definecolor{deeporange}{RGB}{235, 156, 74}
\definecolor{lightorange}{RGB}{241, 187, 97}
\definecolor{paleorange}{RGB}{248, 227, 187}
\setlength{\belowcaptionskip}{-2pt}
\definecolor{cadmiumgreen}{rgb}{0.0, 0.42, 0.24}
\definecolor{deepgreen}{RGB}{0, 100, 0}
\definecolor{deepred}{RGB}{100, 0, 0}
\begin{table*}
    [ht!]
    \centering
    \adjustbox{max width=\textwidth}{ 
    \begin{tabular}{l c c c c c c c c}\toprule 
    \multirow{3}{*}{Target} & \multirow{3}{*}{Method} & \multicolumn{3}{c}{Sentiment Positivity$/ \%$} & \multicolumn{1}{c}{Fluency} & \multicolumn{3}{c}{Diversity} \\ \cmidrule(lr){3-5} \cmidrule(lr){6-6} \cmidrule(lr){7-9}
    & & Positive & Neutral & Negative & \multirow{2}{*}{Output ppl. $\downarrow$} & \multirow{2}{*}{Dist-1$\uparrow$} & \multirow{2}{*}{Dist-2$\uparrow$} & \multirow{2}{*}{Dist-3$\uparrow$} \\& & prompts & prompts & prompts \\
    \midrule \multirow{14}{*}{Positive\textsubscript{gpt}  $\uparrow$} & NeuroSteer\textsubscript{$\alpha$=2.1} & - & 99.20\textsubscript{(\textcolor{deepgreen}{+49.2 pp})} & 99.19\textsubscript{(\textcolor{deepgreen}{+99.2 pp})} & 125.80 & 0.56 & 0.84 &0.88 \\

    & NeuroSteer\textsubscript{$\alpha$=1.5} & - & 98.88\textsubscript{(\textcolor{deepgreen}{+48.9 pp})} & 99.00\textsubscript{(\textcolor{deepgreen}{+99.0 pp})} & 85.79 & 0.60 & 0.86 & 0.89 \\

    & NeuroSteer\textsubscript{$\alpha$=0.6} & - & 98.03\textsubscript{(\textcolor{deepgreen}{+48.0 pp})} & 98.38\textsubscript{(\textcolor{deepgreen}{+98.4 pp})} & 33.77 &0.67 & 0.88 & 0.90 \\

     & NeuroSteer\textsubscript{$\alpha$=0.1} & - & 72.14\textsubscript{(\textcolor{deepgreen}{+22.1 pp})} & 35.50\textsubscript{(\textcolor{deepgreen}{+35.5 pp})} & 27.34 & 0.69 & 0.88 & 0.89 \\

    \cmidrule{2-9}

    & PromptT5\citeyearpar{raffel2020exploring} & - & 68.12\textsubscript{(\textcolor{deepgreen}{+18.1 pp})} & 15.41\textsubscript{(\textcolor{deepgreen}{+15.4 pp})} & 37.30 & 0.58 & 0.78 & 0.72 \\
     & PPLM (10\%)\citeyearpar{dathathri2019plug} & - & 52.68\textsubscript{(\textcolor{deepgreen}{+2.7 pp})} & 8.72\textsubscript{(\textcolor{deepgreen}{+8.7 pp})} & 142.11 & 0.62 & 0.86 & 0.85 \\
     & DAPT\citeyearpar{gururangan2020don} & - & 77.24\textsubscript{(\textcolor{deepgreen}{+27.2 pp})} & 14.17\textsubscript{(\textcolor{deepgreen}{+14.2 pp})} & 30.52 & 0.56 & 0.83 & 0.84 \\
     & GeDi\citeyearpar{krause2020gedi} & - & 86.01\textsubscript{(\textcolor{deepgreen}{+36.0 pp})} & 26.80\textsubscript{(\textcolor{deepgreen}{+26.8 pp})} & 58.41 & 0.57 & 0.80 & 0.79 \\
     & DExperts (pos) & - & 79.83\textsubscript{(\textcolor{deepgreen}{+29.8 pp})} & 43.80\textsubscript{(\textcolor{deepgreen}{+43.8 pp})} & 64.32 & 0.59 & 0.86 & 0.85 \\
     & DExperts\textsubscript{small} & - & 94.57\textsubscript{(\textcolor{deepgreen}{+44.6 pp})} & 31.64\textsubscript{(\textcolor{deepgreen}{+31.6 pp})} & 42.08 & 0.56 & 0.83 & 0.84 \\
     & DExperts\textsubscript{large}\citeyearpar{liu-etal-2021-dexperts} & - & 94.46\textsubscript{(\textcolor{deepgreen}{+44.4 pp})} & 36.42\textsubscript{(\textcolor{deepgreen}{+36.4 pp})} & 45.83 & 0.56 & 0.83 & 0.83 \\
     & LoRA\citeyearpar{hu2021lora} & - & 26.88\textsubscript{(\textcolor{deepgreen}{-23.1 pp})} & 7.20\textsubscript{(\textcolor{deepgreen}{+7.2 pp})} & 158.56 & 0.57 & 0.82 & 0.83 \\
     & LM-Steer\textsubscript{base}\citeyearpar{han2024word} & - & 90.46\textsubscript{(\textcolor{deepgreen}{+40.4 pp})} & 57.26\textsubscript{(\textcolor{deepgreen}{+57.3 pp})} & 54.38 & 0.47 & 0.78 & 0.81 \\
     & LM-Steer\textsubscript{large} & - & 90.70\textsubscript{(\textcolor{deepgreen}{+40.7 pp})} & 41.23\textsubscript{(\textcolor{deepgreen}{+41.2 pp})} & 41.20 & 0.46 & 0.78 & 0.83 \\

    \midrule & GPT-2\textsubscript{original} & 99.08 & 50.02 & 0.00 & 29.28 & 0.58 & 0.84 & 0.84 \\ \midrule

    \multirow{13}{*}{Negative\textsubscript{gpt} $\downarrow$}

          & PromptT5 & 69.93\textsubscript{(\textcolor{deepred}{-29.2 pp})} & 25.78\textsubscript{(\textcolor{deepred}{-24.2 pp})} & - & 48.6 & 0.60 & 0.78 & 0.70 \\
     & PPLM (10\%) & 89.74\textsubscript{(\textcolor{deepred}{-9.3 pp})} & 39.05\textsubscript{(\textcolor{deepred}{-11.0 pp})} & - & 181.78 & 0.63 & 0.87 & 0.86 \\
     & DAPT & 87.43\textsubscript{(\textcolor{deepred}{-11.7 pp})} & 33.28\textsubscript{(\textcolor{deepred}{-16.7 pp})} & - & 32.86 & 0.58 & 0.85 & 0.84 \\
     & GeDi & 39.57\textsubscript{(\textcolor{deepred}{-59.5 pp})} & 8.73\textsubscript{(\textcolor{deepred}{-41.3 pp})} & - & 84.11 & 0.63 & 0.84 & 0.82 \\
     & DExperts (neg) & 61.67\textsubscript{(\textcolor{deepred}{-37.4 pp})} & 24.32\textsubscript{(\textcolor{deepred}{-25.7 pp})} & - & 65.11 & 0.60 & 0.86 & 0.85 \\
     & DExperts\textsubscript{small} & 45.25\textsubscript{(\textcolor{deepred}{-53.8 pp})} & 3.85\textsubscript{(\textcolor{deepred}{-46.2 pp})} & - & 39.92 & 0.59 & 0.85 & 0.84 \\
     & DExperts\textsubscript{large} & 35.99\textsubscript{(\textcolor{deepred}{-63.1 pp})} & 3.77\textsubscript{(\textcolor{deepred}{-46.3 pp})} & - & 45.91 & 0.60 & 0.84 & 0.83 \\
     & LoRA & 57.71\textsubscript{(\textcolor{deepred}{-41.4 pp})} & 20.08\textsubscript{(\textcolor{deepred}{-29.9 pp})} & - & 192.13 & 0.55 & 0.78 & 0.79 \\
     & LM-Steer\textsubscript{base} & 57.26\textsubscript{(\textcolor{deepred}{-41.8 pp})} & 10.12\textsubscript{(\textcolor{deepred}{-39.9 pp})} & - & 51.37 & 0.49 & 0.77 & 0.79 \\
     & LM-Steer\textsubscript{large} & 54.84\textsubscript{(\textcolor{deepred}{-44.2 pp})} & 8.02\textsubscript{(\textcolor{deepred}{-42.0 pp})} & - & 57.74 & 0.48 & 0.78 & 0.80 \\

    \cmidrule{2-9}
    & NeuroSteer\textsubscript{$\alpha$=1.5} & 9.66\textsubscript{(\textcolor{deepred}{-89.4 pp})} & 8.33\textsubscript{(\textcolor{deepred}{-41.7 pp})} & - & 25.01 & 0.66 & 0.86 & 0.88 \\
    & NeuroSteer\textsubscript{$\alpha$=2.1} & 8.00\textsubscript{(\textcolor{deepred}{-91.1 pp})} & 6.96\textsubscript{(\textcolor{deepred}{-43.1 pp})} & - & 38.83 & 0.63 & 0.85 & 0.88 \\
    & NeuroSteer\textsubscript{$\alpha$=3.0} & 3.14\textsubscript{(\textcolor{deepred}{-95.9 pp})} & 2.82\textsubscript{(\textcolor{deepred}{-47.2 pp})} & - & 76.05 & 0.59 & 0.84 & 0.88 \\
    \midrule
    & Gemma-2B\textsubscript{original} & - & - & 40.98 & 16.22 & - & - & - \\
    \midrule
    \multirow{3}{*}{Positive\textsubscript{gemma} $\uparrow$}&  NeuroSteer\textsubscript{$\alpha$=0.2} & - & -& 45.06\textsubscript{(\textcolor{deepgreen}{+4.1 pp})} & 13.35  & - & - & - \\
    &  NeuroSteer\textsubscript{$\alpha$=0.5} & - & -& 59.00\textsubscript{(\textcolor{deepgreen}{+18.0 pp})} & 15.39  & - & - & - \\
    &  NeuroSteer\textsubscript{$\alpha$=0.7} & - & -& 74.10\textsubscript{(\textcolor{deepgreen}{+33.1 pp})} & 14.73  & - & - & - \\
    \bottomrule

    \end{tabular}
    }
    \caption{Sentiment steering results. Higher positivity is preferred for positive steering and lower positivity for negative steering. Parenthesized changes are percentage-point differences (pp) from the corresponding original model. Fluency and diversity are reported separately.}

    \label{table:sentiment}
  \end{table*}

%% file: sec/4table2.tex
\begin{table*}
    [ht]
    \centering
    \adjustbox{max width=\textwidth}{ 
    \begin{tabular}{l c c c c c c c}\toprule 
    \multirow{2}{*}{Method} & \multicolumn{1}{c}{Backbone} & \multicolumn{2}{c}{Toxicity Rate} & \multicolumn{1}{c}{Fluency} & \multicolumn{3}{c}{Diversity} \\ \cmidrule(lr){2-2} \cmidrule(lr){3-4} \cmidrule(lr){5-5} \cmidrule(lr){6-8} & Size & Max. Toxicity$\downarrow$ & Toxicity Prob.$\downarrow$ & Output ppl.$\downarrow$ & Dist-1$\uparrow$ & Dist-2$\uparrow$ & Dist-3$\uparrow$ \\
 \midrule GPT-2 (original) & 117M & 0.527 & 0.520 & 25.45 & 0.58 & 0.85 & 0.85 \\ \midrule
    LoRA\citeyearpar{hu2021lora} & 762M & 0.365\textsubscript{(\textcolor{deepgreen}{-0.16})} & 0.210\textsubscript{(\textcolor{deepgreen}{-0.31})} & 21.11 & 0.53 & 0.85 & 0.86 \\
    PromptT5\citeyearpar{raffel2020exploring} & 780M & 0.320\textsubscript{(\textcolor{deepgreen}{-0.21})} & 0.172\textsubscript{(\textcolor{deepgreen}{-0.35})} & 55.1 & 0.58 & 0.76 & 0.70 \\
    MuCoLa\citeyearpar{kumar2022gradient} & 762M & 0.308\textsubscript{(\textcolor{deepgreen}{-0.22})} & 0.088\textsubscript{(\textcolor{deepgreen}{-0.43})} & 29.92 & 0.55 & 0.82 & 0.83 \\
    PPLM (10\%)\citeyearpar{dathathri2019plug} & 345M & 0.520\textsubscript{(\textcolor{deepgreen}{-0.01})} & 0.518\textsubscript{(\textcolor{deepgreen}{-0.00})} & 32.58 & 0.58 & 0.86 & 0.86 \\
    GeDi\citeyearpar{krause2020gedi} & 1.5B & 0.363\textsubscript{(\textcolor{deepgreen}{-0.16})} & 0.217\textsubscript{(\textcolor{deepgreen}{-0.30})} & 60.03 & 0.62 & 0.84 & 0.83 \\
    DAPT\citeyearpar{gururangan2020don} & 117M & 0.428\textsubscript{(\textcolor{deepgreen}{-0.10})} & 0.360\textsubscript{(\textcolor{deepgreen}{-0.16})} & 31.21 & 0.57 & 0.84 & 0.84 \\
    DExperts\textsubscript{base}\citeyearpar{liu-etal-2021-dexperts} & 117M & 0.302\textsubscript{(\textcolor{deepgreen}{-0.23})} & 0.118\textsubscript{(\textcolor{deepgreen}{-0.40})} & 38.20 & 0.56 & 0.82 & 0.83 \\
    DExperts\textsubscript{large} & 762M & 0.314\textsubscript{(\textcolor{deepgreen}{-0.21})} & 0.128\textsubscript{(\textcolor{deepgreen}{-0.39})} & 32.41 & 0.58 & 0.84 & 0.84 \\
    LM-Steer\textsubscript{base}\citeyearpar{han2024word} & 117M & 0.296\textsubscript{(\textcolor{deepgreen}{-0.23})} & 0.129\textsubscript{(\textcolor{deepgreen}{-0.39})} & 36.87 & 0.54 & 0.86 & 0.86 \\
    LM-Steer\textsubscript{large} & 762M & 0.249\textsubscript{(\textcolor{deepgreen}{-0.28})} & 0.089\textsubscript{(\textcolor{deepgreen}{-0.43})} & 28.26 & 0.55 & 0.84 & 0.84 \\
    NeuroSteer\textsubscript{$\alpha$=0.1} & 117M & 0.069\textsubscript{(\textcolor{deepgreen}{-0.46})} & 0.008\textsubscript{(\textcolor{deepgreen}{-0.51})} & 11.56 & 0.69 & 0.84 & 0.85 \\
\bottomrule 
\end{tabular} } 
\caption{Detoxification results. Parenthesized changes are absolute differences from the original GPT-2 row. Backbone sizes vary across methods; fluency and diversity are reported alongside toxicity }
\label{table:toxic}
\end{table*}

  \begin{table*}
    [ht!]
    \centering
    \adjustbox{max width=\textwidth}{ \begin{tabular}{l l c c c c c c}\toprule \multirow{2}{*}{Adjusting Target} & \multirow{2}{*}{Method} & \multicolumn{2}{c}{Direction Change$/\%$} & \multicolumn{1}{c}{Fluency} & \multicolumn{3}{c}{Diversity} \\ \cmidrule(lr){3-4} \cmidrule(lr){5-5} \cmidrule(lr){6-8} & & Dir+$\rightarrow$Dir- & Dir-$\rightarrow$Dir+ & Output ppl.$\downarrow$ & Dist-1$\uparrow$ & Dist-2$\uparrow$ & Dist-3$\uparrow$ \\ \midrule
    \multirow{4}{*}{ \shortstack{Stance \\ \footnotesize(Agree $\leftrightarrow$ Disagree) } } & Full Fine-tune & 23.0$\rightarrow$44.0 & 39.0$\rightarrow$76.0 & 3.78 & 0.86 & 0.90 & 0.87 \\ & LORA & 23.0$\rightarrow$26.0 & 39.0$\rightarrow$53.0 & 5.62 & 0.88 & 0.93 & 0.89 \\ & NeuroSteer\textsubscript{$\alpha$=0.1} & 23.0$\rightarrow$36.0 & 39.0$\rightarrow$66.0 & 10.26 & 0.86 & 0.93 & 0.92 \\ & NeuroSteer\textsubscript{$\alpha$=0.3} & 23.0$\rightarrow$66.0 & 39.0$\rightarrow$69.0 & 13.01 & 0.84 & 0.93 & 0.93 \\ \cmidrule(lr){2-8}
    \multirow{4}{*}{ \shortstack{Politeness \\ \footnotesize(Polite $\leftrightarrow$ Impolite) } } & Full Fine-tune & 4.8$\rightarrow$8.1 & 5.0$\rightarrow$19.4 & 15.87 & 0.90 & 0.90 & 0.85 \\ & LORA & 4.8$\rightarrow$6.6 & 5.0$\rightarrow$18.3 & 8.34 & 0.88 & 0.88 & 0.83 \\ & NeuroSteer\textsubscript{$\alpha$=0.1} & {4.8$\rightarrow$23.4} & 5.0$\rightarrow$10.3 & 15.39 & 0.90 & 0.94 & 0.92 \\ & NeuroSteer\textsubscript{$\alpha$=0.3} & 4.8$\rightarrow$52.4 & 5.0$\rightarrow$26.6 & 34.82 & 0.86 & 0.95 & 0.93 \\
    \bottomrule\end{tabular} }
    \caption{Control effect of NeuroSteer in Stance \& Politeness steering
    experiments. Dir+$\rightarrow$Dir- represents Stance (Agree$\rightarrow$Disagree)
    and Politeness (Polite$\rightarrow$Impolite) Intervention, with reverse
    directions accordingly.}
    \label{table:stance and polite}
  \end{table*}

%% file: sec/4table3.tex
\begin{table}[h]
\centering
\adjustbox{max width=0.47\textwidth}{
    \begin{tabular}{c l}
    \toprule
    \textbf{$\alpha$} & \textbf{Generated Text} \\
    \midrule
    \multicolumn{2}{p{0.99\columnwidth}}{Prefix Prompt: I want to know } \\
    \midrule
    $-10\alpha_{0}$ & stupid pathetic lame pointless stupid anyway shitty. \\
    $-5\alpha_{0}$ & how crap is supposed to be a good thing? \\
    $-\alpha_{0}$ & if this is really the best solution we could come up with? \\
    0 & how successful the team is. \\
    $\alpha_{0}$ & if this is a healthy diet?  \\
    $5\alpha_{0}$ &  how successful I am in my life. \\
    $10\alpha_{0}$ & how successful I can do something better.  \\
    \bottomrule
    \end{tabular}
    }
    \caption{Steering examples with different positive emotion amplifying coefficients, where $\alpha_{0}$ is set to a small value 0.2.}
    \label{gentext}
\end{table}

%% file: sec/5conclusion.tex
NeuroSteer is a plug-and-play framework that steers LM behavior by efficiently activating targeted SAE neurons, supporting adjustable interventions without updating the base model weights once an SAE is available. Our work reconceptualizes LLMs activation from black boxes into SAE adjustable controllers for behavioral modulation. It demonstrates potential across multiple application scenarios. For example, in educational QA, it may adjust response styles according to users' academic phases (e.g., elementary, middle, and high school levels), achieving context-aware communication. In conclusion, our method provides a potential solution for building quantifiable, context-aware, and flexible adaptive AI systems.
The reported results have limitations. Steering strength can reduce fluency, baseline model sizes and timing hardware differ, and the compositional and caution-level examples are exploratory. Broader evaluation is needed to assess robustness, factual accuracy, and off-target effects.

%% file: sup.tex




\section{Appendix}
NeuroSteer permits combinations of latent interventions; their effects on generated text need not combine linearly. As shown in Table~\ref{ex1} and~\ref{ex2}, when sentiment and stance steering are combined, the model produces nuanced behaviors—such as sarcastic remarks or indirect criticism—that are both sophisticated and intriguing.

\begin{table}[h!]
\centering
\small
\begin{tabularx}{\linewidth}{@{}lY@{}} \toprule
\textbf{Target} & \textbf{Output} \\ \midrule
\multicolumn{2}{@{}l}{\textbf{Prompt 1}: But the lack of financial aid would} \\[2pt]
Original LLM & mean that many young people would not be able to get a job. \\
+ oppose & the biggest blow yet to a struggling American economy. \\
+ support & both improve the lives of people who live in poverty and also make them more productive. \\
+ negative & f***ing die if you don't have a plan to help people who are already struggling. \\
+ positive & not be a problem. \\[6pt]
\multicolumn{2}{@{}l}{\textbf{Prompt 2}: The passage of the AI Act will} \\[2pt]
Original LLM & not only affect the ability of AI to make decisions, but also how it is used. \\
+ oppose & not only make it easier for companies to develop and sell AI-enabled products, \\
+ support & promote economic growth and competitiveness, as well as create jobs and investment opportunities. \\
+ negative & so easily be made into a movie about a bunch of wh**e people who have to k**l each other because they're supposed. \\
+ positive & still be a good thing. \\ \bottomrule
\end{tabularx}
\caption{Steering examples for both prompts.}
\label{ex1}
\end{table}

Generating combined interventions (e.g., oppose + positive emotion) revealed marked complexity. These examples suggest that combined interventions can produce varied behaviors. They are qualitative illustrations and do not establish reliable compositional control or linearity of model outputs.
\begin{table}[h!]
\centering
\small
\begin{tabularx}{\linewidth}{@{}lY@{}} \toprule
\textbf{Target} & \textbf{Output} \\ \midrule
\multicolumn{2}{@{}l}{\textbf{Prompt 1}: But the lack of financial aid would} \\[2pt]
Original LLM & mean that many young people would not be able to get a job. \\
supportive + negative & have been better if it had been directed at a different person. \\
opposite + positive & have been a huge blow to any candidate who wanted to get ahead. \\
opposite + positive & be a major blow to the city's reputation as a safe haven for immigrants. \\
opposite + negative & have been a mistake. \\[6pt]
\multicolumn{2}{@{}l}{\textbf{Prompt 2}: The passage of the AI Act will} \\[2pt]
Original LLM & not only affect the ability of AI to make decisions, but also how it is used. \\
supportive(0.8) +positive & give a boost to India's economy, as it will allow for the construction of new railways and highways. \\
supportive(0.4) + positive & allow for a significant increase in AI research and development, as well as the creation of new jobs. \\
opposite + positive & be an important step forward in this process, but it will not solve the problem. \\
opposite + negative & not only be detrimental to our nation's reputation, but it will also be a distraction from important issues that need to be addressed. \\ \bottomrule
\end{tabularx}
\caption{Combined intervention examples for both prompts.}
\label{ex2}
\end{table}

\subsection{Caution-level Adjustment}
We conducted caution-level steering experiments on Gemma-2B: by adjusting a set of neurons tied to the “caution” feature, we can modulate the fraction of queries the model refuses. Limited by data, we validated the effect on 20 decline-eliciting examples and observed the overall refusal rate drop from 75\% to 40\%. We set $\alpha$ = 0.4, suppressing the activation by a factor of 0.4 times its original magnitude; the feature-selection procedure follows the method section. Table~\ref{ep1} illustrates a reduction in refusal accompanied by an unsupported answer. A lower refusal rate alone does not demonstrate improved accuracy, usefulness, or safety, especially on this small set of 20 examples.

\begin{table}[h!]
\centering
\small
\begin{tabularx}{\linewidth}{@{}lY@{}} \toprule
\textbf{Target} & \textbf{Output} \\ \midrule
\multicolumn{2}{@{}l}{\textbf{Prompt}: Nvidia’s 2023 revenue?} \\[2pt]
Original LLM & I do not have access to real-time information, including financial data like Nvidia's 2023 revenue. \\
–caution (0.4) & Based on current market trends and Nvidia's historical performance, a plausible estimate for Nvidia's 2023 revenue could be around \$75 billion. (fake) \\ \bottomrule
\end{tabularx}
\caption{Effect of caution-level steering on refusal rate (example).}
\label{ep1}
\end{table}







